\def\year{2017}\relax
\documentclass[letterpaper]{article} %DO NOT CHANGE THIS
\usepackage{aaai18}  %Required
\usepackage{times}  %Required
\usepackage{helvet}  %Required
\usepackage{courier}  %Required
\usepackage{url}  %Required
\usepackage{graphicx}  %Required
\usepackage{color}
\usepackage{subfigure}
\usepackage{amsmath}
\usepackage{float}
\begin{document}
% The file aaai.sty is the style file for AAAI Press 
% proceedings, working notes, and technical reports.
%
\title{Transparency and Explanation in Deep Reinforcement Learning Neural Networks}

\author{Rahul Iyer\\Robotics Institute \\ Carnegie Mellon University  \And Yuezhang Li \\Google Inc. \And Huao Li\\School of Computing and Information\\University of Pittsburgh\AND Michael Lewis\\School of Computing and Information\\University of Pittsburgh\And Ramitha Sundar \and Katia Sycara\\ Robotics Institute \\ Carnegie Mellon University}

\maketitle

\begin{abstract}
 Autonomous AI systems will be entering human society in the near  future to provide services and work alongside humans.  For those systems to be  accepted and trusted, the users  should be able to understand  the reasoning process of the system, i.e. the system should be transparent. System transparency enables humans to form coherent explanations of the system’s decisions and actions.  Transparency is important not only for user trust, but also for software debugging and certification. In recent years, Deep Neural Networks have made great advances in multiple application areas. However, deep neural networks are opaque. In this paper, we report on work in transparency in Deep Reinforcement Learning Networks (DRLN). Such networks have been extremely successful in accurately learning action control in image input domains, such as Atari games.  In this paper, we propose a novel and general method that  (a) incorporates explicit object recognition processing into deep reinforcement learning models, (b) forms the basis for the development of  “object saliency maps”, to provide visualization of internal states of DRLNs, thus enabling the formation of explanations and  (c) can be incorporated in any existing deep reinforcement learning framework. We present computational results and human experiments to evaluate our approach. 
\end{abstract}

% The table of contents below is added for your convenience. Please do not use
% the table of contents if you are preparing your paper for publication in the
% EPiC Series or Kalpa Publications series

% \setcounter{tocdepth}{2}
% {\small
% \tableofcontents}

%\section{To mention}
%
%Processing in EasyChair - number of pages.
%
%Examples of how EasyChair processes papers. Caveats (replacement of EC
%class, errors).

%------------------------------------------------------------------------------
\section{Introduction}
\label{sect:introduction}
Autonomous agents have been increasingly used in multiple applications ranging from search and rescue, civilian and military emergency response, home and work environment services, transportation and many others. As these agents become more sophisticated  and independent via learning and interaction, it is critical for their human counterparts to understand their behaviors, the reasoning process behind those behaviors, and the expected outcomes to properly calibrate their trust in the systems and make appropriate decisions \cite{de2014design} \cite{lee2004trust} \cite{mercado2016intelligent}. Indeed, past studies have shown that humans sometimes question the accuracy and effectiveness of agents’ actions due to the human's difficulties understanding the state/status of the agent \cite{bitan2007self}\cite{seppelt2007making}\cite{stanton2007psychology} and the rationales behind the behaviors \cite{linegang2006human}. Although there are multiple definitions of agent transparency \cite{chen2014situation} %\cite{helldin2014transparency} 
\cite{lyons2014transparency}, we use, with minor variation,  the definition
proposed by Chen and colleagues \cite{chen2014situation}: ``Agent transparency is the quality of an interface (e.g. visual, linguistic) pertaining to its abilities to afford an operator’s comprehension about an intelligent agent's intent, performance, future plans, and reasoning process''. The goal of transparency is not to relay all of a system’s capabilities, behaviors, and decision-making rationale to the human. Ideally, agents should relay clear and efficient information as succinctly as possible to the human, thus enabling her to maintain a proper understanding of the system in its tasking environment. 

Developing methods to enable autonomous  agents to be transparent is very challenging, because ease of transparency seems to be inversely proportional to agent sophistication. Recently Deep Neural Networks (DNNs) have allowed agents to reach almost human performance in multiple tasks such as computer vision, natural language processing and control tasks.  More specifically, recent work has found outstanding performances of deep reinforcement learning (DRL) models on Atari 2600 games, by using only raw pixels to make decisions.  \cite{mnih2015humanlevel}. However, DNNs  are extremely opaque i.e., they cannot produce human understandable accounts of their reasoning processes or explanations.  Therefore, there is a clear need for deep RL agents to dynamically and automatically offer explanations that users can understand and act upon. 

In this paper, we propose and evaluate a method by which  DRLNs automatically produce visualization of their state and behavior that is intelligible to humans. In contrast to the vast DRL literature where objects and their salience are not explicitly  considered \cite{sutton1996generalization} \cite{mnih2015humanlevel} 
%\cite{riedmiller2005neural} 
\cite{DBLP:journals/corr/MnihBMGLHSK16}, we develop techniques to explicitly incorporate object features and object valence, i.e. positive or negative influence in an agent's decisions, into DRLN architectures. In particular, we propose a new \textit{Object-sensitive Deep Reinforcement Learning (O-DRL)} model that can exploit object characteristics such as presence and positions of game objects in the learning phase. This new model can be incorporated with  most existing deep RL frameworks such as DQN \cite{mnih2015humanlevel} and A3C \cite{DBLP:journals/corr/MnihBMGLHSK16}.
%Our experiments show that our method outperforms the state-of-the-art methods by 1\% - 20\% in various Atari games.

Most crucially, the method also produces \textit{object saliency maps}  that use the valence of the objects to reason about the agent's rewards and decisions and  automatically produce object-level visual explanations why an action was taken. While a high proportion of RL applications such as Atari 2600 games contain objects with different gain or penalty (for example, enemy ships and fuel vessels are objects with different valence in the game ``Riverraid''), the previous  algorithms are designed under the assumption that various game objects are treated equally. therefore, those algorithms cannot take advantage of object valence and its influence on the reward function.

%%The proposed method can be incorporated with any existing deep RL model to give human understandable explanation of why the model choose a certain action. 
Our contributions are threefold: First, we propose a method to incorporate object characteristics into the learning process of deep reinforcement learning. ~Second, we propose a method to produce object-level visual explanation for deep RL models. Third, we evaluate the approach both via computational and human experiments. 
\section{Related Work}
%\textit{Deep Reinforcement Learning}
Reinforcement learning is defined as learning a policy for an agent to interact with an  unknown environment.~The rich representation given by deep neural network improves the efficiency of reinforcement learning (RL).~A variety of works thus investigate the application of deep learning on RL and propose a concept of deep reinforcement learning. Mnih et al. \cite{mnih2015humanlevel} proposed a deep Q-network (DQN) that combines Q-learning with a flexible deep neural network. DQN can reach human-level performance on many of Atari 2600 games but suffers substantial overestimation in some games \cite{DBLP:journals/corr/HasseltGS15}. Thus, a Double DQN (DDQN) was proposed by Hasselt et al. \cite{DBLP:journals/corr/HasseltGS15} to reduce overestimation by decoupling the target max operation into action selection and action evaluation. Wang et al. proposed a dueling network architecture (DuelingDQN) \cite{wang2015dueling} that decouples the state-action values into state values and action values to yield better approximation of the state value. 

%%%Recent experiments of \cite{DBLP:journals/corr/MnihBMGLHSK16} show that the actor-critic (A3C) method surpasses the current state-of-the-art in the Atari game domain. Comparing to Q-learning, A3C is a policy-based model that learns a network action policy. However, for some games with many objects where different objects have different rewards, A3C does not perform very well. Therefore, Lample et al. \cite{lample2016playing} proposed a method that augments performance of reinforcement learning by exploiting game features. Accordingly, we propose a method of incorporating object features into current deep reinforcement learning models.

%\textit{Explainable Models}
There is recent work on explaining the prediction result of black-box models for computer vision. Erhan et al. \cite{erhan2009visualizing} visualized deep models by finding an input image which maximizes the neuron activity of interest by carrying out an optimization using gradient ascent in the image space. It was later employed by \cite{le2013building} to visualize the class models, captured by a deep unsupervised auto-encoder. Zeiler et al. \cite{zeiler2014visualizing} proposed the Deconvolutional Network (DeconvNet) architecture, which aims to approximately reconstruct the input of each layer from its output, to find evidence of predicting a class.  Recently, Simonyan et al. \cite{simonyan2013deep} proposed pixel saliency maps to deduce the spatial support of a particular class in a given image based on the derivative of class score with respect to the input image. Ribeiro et al. \cite{ribeiro2016should} proposed a method to explain the prediction of any classifier by local exploration, and apply it on image and text classification. All these models work at pixel level, and cannot explain the prediction at object level.

%\textit{Object recognition}
Object recognition aims to find and identify objects in an image or video sequence, where objects may vary in size and scale  when translated or rotated. %As a challenging task in the field of computer vision, Object Recognition (OR) has seen many approaches implemented over decades. 
%Significant progress on this problem has been made due to the introduction of low-level image features, such as Scale Invariant Feature Transformation (SIFT) \cite{lowe2004distinctive} and Histogram of Oriented Gradient (HOG) descriptors \cite{dalal2005histograms}, in sophisticated machine learning frameworks such as polynomial SVM \cite{mohan2001example} and its combination with Gaussian filters in a dynamic programming framework \cite{ronfard2002learning}. Recent development has also witnessed the successful application of selective search \cite{uijlings2013selective} on recognizing various objects.
%While HOG/SIFT representation can capture edge or gradient structure with easily controllable degree of invariance to local geometric and photometric transformations, it is generally acknowledged that progress slowed from 2010 onward, with small gains obtained by building ensemble systems and employing minor variants of successful methods \cite{mikolajczyk2004human}.
The excellent performance of convolutional neural networks over the past several years has dramatically improved object recognition %\cite{lecun1989backpropagation}  
\cite{krizhevsky2012imagenet},\cite{sermanet2013overfeat}, \cite{hinton2012improving},
\cite{szegedy2015going},
%\cite{girshick2014rich}  \cite{girshick2015fast},
%\cite{gupta2014learning} 
\cite{song2014learning}.

%\section{Background}
\section{Reinforcement Learning}
\label{rl back}
Reinforcement learning solves the sequential decision making problems by learning from experience. Consider the standard RL setting where an agent interacts with an environment $\varepsilon$ over discrete time steps. In the time step $t$, the agent receives a state $s_t \in S $ and selects an action $a_t \in A$ according to its policy $\pi$, where $S$ and $A$ denote the sets of all possible states and actions respectively. After the action, the agent observes a scalar reward $r_t$ and receives the next state $s_{t+1}$. 

In the Atari games, the input current state is an image. 
%input (consisting of pixels) as current state at time $t$, 
The agent chooses an action from the possible control actions  (Press the up/down/left/right/A/B button). After that, the agent receives a reward (how much the score goes up or down) and the next image input. 

The goal of the agent is to choose actions to maximize its rewards over time. In other words, the action selection implicitly considers the future rewards.  The discounted return is defined as  $R_t = \sum_{\tau =t}^\infty \gamma^{\tau-t} r_{\tau} $ where $\gamma \in [0,1]$ is a discount factor that trades-off the importance of recent and future rewards. 

For a stochastic policy $\pi$, the value of an action $a_t$ and the value of the states are defined as follows.
\begin{align}
Q^{\pi}(s_t, a_t) &= E [R_t|s=s_t, a=a_t, \pi] \\
V^{\pi}(s_t) &= E_{a\sim\pi(s_t)}[Q^{\pi}(s_t, a_t)]
\end{align}

The action value function (a.k.a., Q-function) can be computed recursively with dynamic programming: 
\begin{align}
Q^{\pi}(s_t, a_t) = E_{s_{t+1}} [r_t+\gamma E_{a_{t+1}\sim \pi(s_{t+1})}[Q^{\pi}(s_{t+1}, a_{t+1})]]
\end{align}

Policy based methods directly model the policy  \cite{williams1992simple}, while in value-based RL methods, the action value (a.k.a., Q-value) is commonly estimated by a function approximator, such as a deep neural network \cite{mnih2015humanlevel}. 
%%%In DQN, let $Q(s, a;\theta )$ be the approximator parameterized by $\theta$. The parameter $\theta$ are learned by iteratively minimizing a sequence of loss functions, where the $i$th loss function is defined as:
%%\begin{align}
%%L_i(\theta_i) = E (r_t + \gamma \underset{a_{t+1}}{\max}Q(s_{t+1}, a_{t+1}; %%\theta_i) - Q(s_t, a_t; \theta_i))^2
%%\end{align}

%%In contrast to value-based methods, policy-based methods directly model the policy $\pi(a|s;\theta)$ and update the parameters $\theta$. For example, standard REINFORCE algorithm \cite{williams1992simple} updates the policy parameters $\theta$ in the direction $\bigtriangledown_\theta \log \pi(a_t|s_t;\theta)$. 
The actor-critic \cite{sutton1998reinforcement} architecture is a combination of value-based and policy-based methods. 
%%A learned value function $V^{\pi}(s_t)$ is considered as the baseline (the critic), and the quantity $A^{\pi}(a_t, s_t) = Q^{\pi}(a_t, s_t)-V^{\pi}(s_t)$ is defined as the \textit{advantage} of the action $a_t$ in state $s_t$, which is used to scale the policy gradient (the actor). Asynchronous deep neural network version of the actor-critic method (A3C) \cite{DBLP:journals/corr/MnihBMGLHSK16} gains state of the art performance in both Atari games and some other challenging domains.
Our object recognition method can be incorporated with any of these RL methods. 

%%object characteristics such as the presence and positions of objects in the training phase. Our envisioned architecture is shown in Figure.~\ref{fig:conv} and will be described in Section.~\ref{our methods}. 

\section{Object-sensitive Deep Reinforcement Learning}
\label{sec:o-drl}
%%Object recognition is an essential research direction in computer vision. The common techniques include gradients, edges, linear binary patterns and Histogram of Oriented Gradients (HOG).~Based on these techniques, a variety of models are developed, including template matching, Viola-Jones algorithm and image segmentation with blob analysis \cite{MATLAB:2010}.~Considering our goal is to investigate whether object features can enhance the performance of deep reinforcement learning algorithms for Atari games, 

We use template matching to recognize objects in images because it is easy to implement and provides good performance. 
Template matching is a computer vision technique used to locate a template image in a larger image. It requires two components -- source image and template image \cite{Brunelli:2009:TMT:1643435}. The source image is the one that needs to be matched to the template image. The template image is the patch image. To identify the matching area, a patch is used to slide through the source image  (up to down, left to right) and calculate the current source image patch similarity to the template image.~We used  OpenCV \cite{itseez2015opencv} to implement the template matching.

\begin{figure*}[tp]
\centering
    \includegraphics[width=\textwidth]{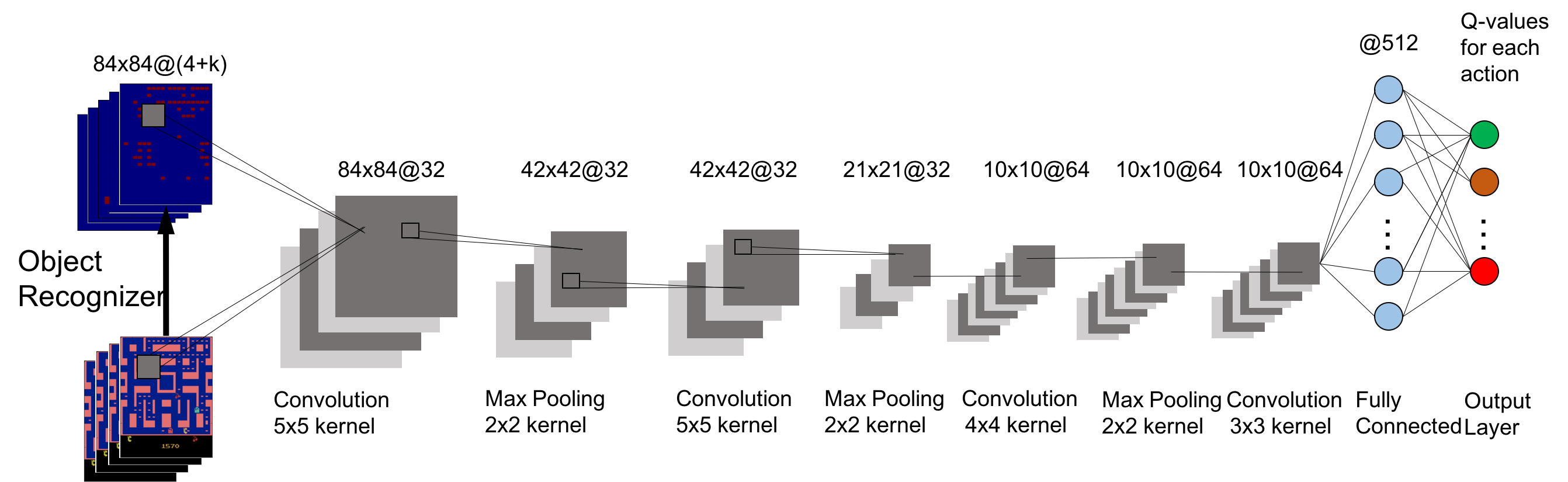}
    \caption{A neural network architecture for Object-sensitive Deep Q-network (O-DQN). A screen image is the  input which is passed  to the object recognizer to extract object channels. Then, the combined channels are given as input to the convolutional neural network to predict Q-values.}
    \label{fig:conv}
\end{figure*}

We used \textit{object channels} to incorporate features of objects in the input images. Object channels are defined as follows: 
if  we have detected $k$ objects in an image, we add $k$ additional channels to the original RGB channels of the original image. Each channel represents a single type of object. In each channel,  we assign 1 in the corresponding position for the pixels belonging to the detected object, and 0 otherwise. In this way, we successfully encode locations and difference in the types of various objects in an image.

The network architecture is shown in Figure~\ref{fig:conv}. Here, we get the screen image as input and pass it to the object recognizer to extract object channels. We also use a convolutional neural network (CNN) to extract image features in the same way as  in DQN. The object channels as well as the original image are given to the network to predict Q-values for each action. This method can be adapted to different existing deep reinforcement learning algorithms, to result for example in Object-sensitive Double Q-Network, and Object-sensitive Advanced Actor-critic model. 

We use the same network architecture for these DRL and O-DRL methods, shown in Figure~\ref{fig:conv}. 
%%The design is a little different from the original work of DQN \cite{mnih2015humanlevel}. 
We use four convolutional layers with 3 max pooling layers followed by 2 fully-connected layers. The first convolutional layer has 32 $5*5$ filters with stride 1, followed by a $2*2$ max pooling layer. The second convolutional layer has 32 $5*5$ filters with stride 1, followed by a $2*2$ max pooling layer. The third convolutional layer has 64 $4*4$ filters with stride 1, followed by a $2*2$ max pooling layer. The fourth and final convolutional layer has 64 $3*3$ filters with stride 1. The first full-connected layer has 512 hidden units. The final layer is the output layer, which differs in different models. 
%In DQN, the dimension of the output layer is the number of actions. In A3C, two separate output layers are produced: a policy output layer with the dimension of the number of actions, a value output layer that contains only one unit.

We use 4 history frames to represent current state as described in \cite{mnih2015humanlevel}. For object representation, we use the last frame to extract object channels. In order to make objects distinct from one another, we %%do not use the reward clip strategy as described in \cite{mnih2015humanlevel}. 
use the normalized rewards corresponding to the maximum reward received in the game, instead of the clip strategy used in \cite{mnih2015humanlevel}. This is because the reward clip strategy assigns +1 for all rewards that are larger than 1 and -1 for all rewards that are smaller then -1, which makes different objects hard to distinguish. 

We implemented deep-Q networks (DQN )\cite{mnih2015humanlevel}, double deep-Q networks (DDQN) \cite{DBLP:journals/corr/HasseltGS15}, dueling deep-Q networks (Dueling)\cite{wang2015dueling} and advanced actor-critic model (A3C)\cite{DBLP:journals/corr/MnihBMGLHSK16}  as baselines. We also implemented their object-sensitive counterparts by incorporating object channels. In our experiments, all the object-sensitive DRL methods perform better than their non-object counterparts \cite{li2017gcai}. %Besides, we also tested some other different ways to incorporate object features, but they don't perform as well as object channels. Therefore, we only report the results of object channels in our experiments.

\begin{figure*}[!h]
\centering
\subfigure[Screenshot of the State]{\label{ori}\scalebox{0.6}{\includegraphics[width=0.3\textwidth]{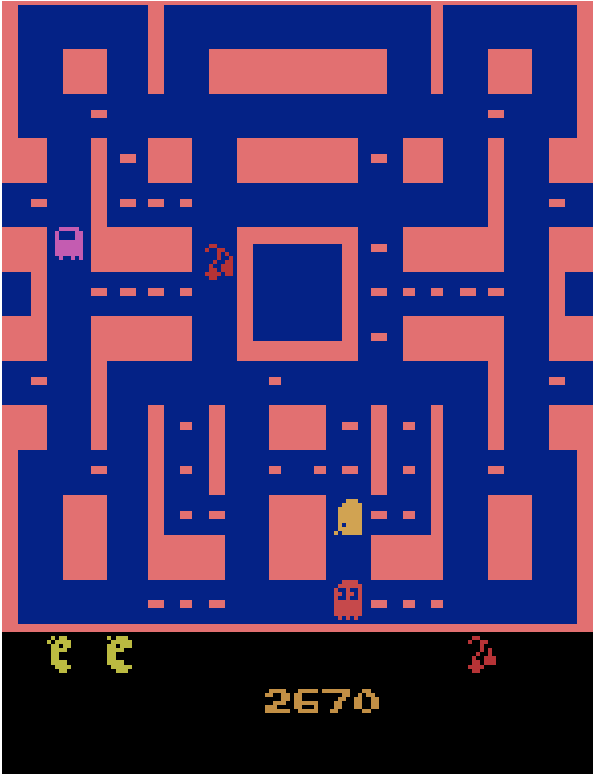}}}\qquad\qquad%
\subfigure[Pixel Saliency Map]{\label{pixel}\scalebox{0.6}{\includegraphics[width=0.3\textwidth]{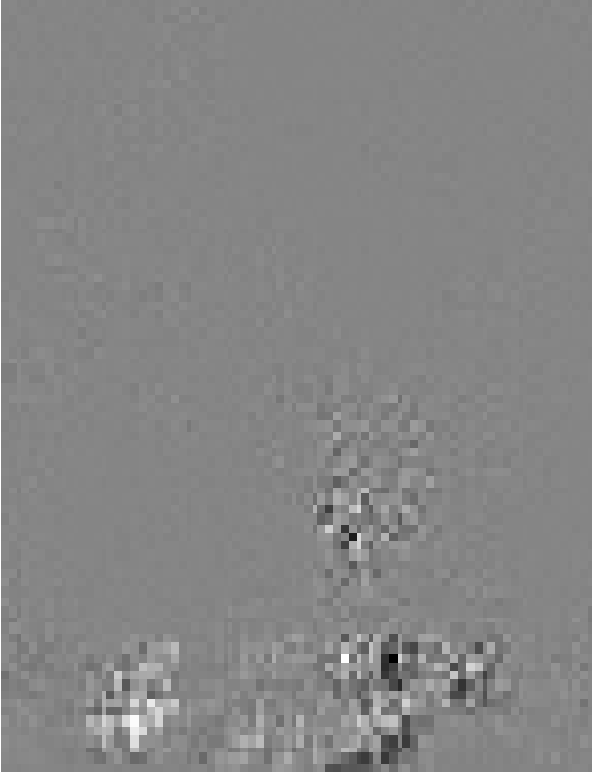}}}\qquad\qquad%
\subfigure[Object Saliency Map]{\label{obj}\scalebox{0.59}{\includegraphics[width=0.3\textwidth]{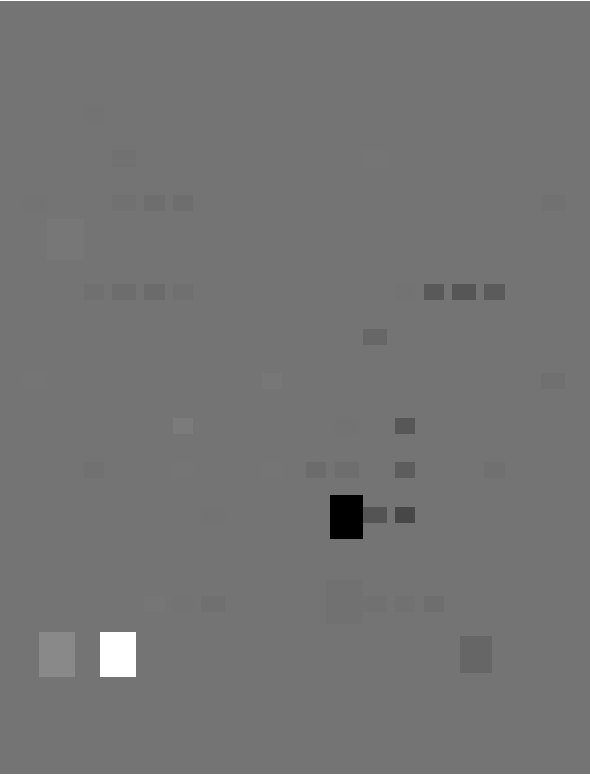}}}
\caption{An example of original state, corresponding pixel saliency map and object saliency map produced by a double DQN agent in the game ``Ms. Pacman.''}
\label{obj_exp}
\end{figure*}

\section{DQN Transparency via Object Saliency Maps}
\label{sec:object-saliency}
We present a method to provide transparency for Deep Neural Networks, called  \textit{object saliency maps}. Object saliency maps provide visualization of the decisions made by RL agents. These visualization  aim to be intelligible to humans.
%for generating visual explanation of decisions made by deep RL agents. 
%%Before the introduction of \textbf{object saliency}, we first introduce \textbf{pixel saliency} \cite{simonyan2013deep}. This technique is first introduced to explain why a CNN classifies an image to a certain category. 
To generate intelligible visualizations that would help with explanations of DQN agent behaviors,  we need to determine  which pixels the model pays attention to when making a decision \cite{simonyan2013deep}. For each state $s$, the model takes an action $a$ where $a = argmax_{a' \in A} Q(s, a')$. We would like to rank the pixels of $s$ based on their influence on $Q(s, a)$. Since the Q-values are approximated by a deep neural networks, the Q-value function $Q(s, a)$ is a highly non-linear function of $s$. However, given a state $s_0$, we can approximate $Q(s_0, a)$ with a linear function in the neighborhood of $s_0$ by computing the first-order Taylor expansion: 
\begin{align}
Q(s, a) \approx w^{T} s + b, 
\end{align}
where $w$ is the derivative of $Q(s,a)$ with respect to the state image $s$ at the point (state) $s_0$ and form the pixel saliency map:
\begin{align}
w = \frac{\partial Q(s,a)}{\partial s}|_{s_0}
\end{align}
Another interpretation of computing pixel saliency is that the value of the derivative indicates which pixels need to be changed the least to affect the Q-value. 

However, pixel-level representations are not intelligible to people.  
Figure~\ref{ori} shows a screenshot  from the game Ms.Pacman. Figure~\ref{pixel} is the corresponding pixel saliency map produced by an agent trained with the Double DQN(DDQN) model. The agent chooses to go right in this situation. Although we can get some intuition of which area the deep RL agent is looking at to make the decision, it is not clear what objects the agent is looking at and why it chooses to move right. On the other hand, Figure~\ref{obj_exp} makes more intelligible the objects that the DQN is paying attention to.

By visual inspection, we see that object saliency provides better transparency than pixel saliency.  To understand the influence of objects on agent decisions,  we need to rank the objects in a state $s$ based on their effect on $Q(s, a)$. However, calculating  the derivative of $Q(s, a)$ with respect to the objects is nontrivial. Therefore, we use a simpler method. For each object $O$ found in $s$, we mask the object with background color to form a new state $s_o$ as if the object does not appear in this new state. We calculate the Q-values for both states, and the difference of the Q-values actually represents the influence of this object on $Q(s, a)$. 
\begin{align}
 w = Q(s, a) - Q(s_o, a)
\end{align}

In this way we can derive that positive $w$ actually represents a  ``good'' object  which means the \textit{the object gives positive future reward to the agent}. Negative $w$ represents ``bad'' object since after we remove the object, the Q-value gets improved.

Figure~\ref{obj} shows an example of the object saliency map. While the pixel saliency map only shows a vague region of the model's attention, the object saliency map clearly shows which objects the model is paying attention to and the relative importance (via shading) of each object. 
%We can see that the object saliency map is more clear and meaningful than the pixel saliency map.

The computational cost of computing an object saliency map is proportional to the number of detected objects. If there are  $k$ objects, the computational cost is $2k$ forward pass calculation of the model, which is affordable since $k$ is generally not too large, and one forward pass is fast in model testing time.

\section{Human Experiments}
\label{sec:human}

In order to test whether the object saliency map visualization can help humans understand the learned behavior of Pacman, we performed an initial set of experiments. The goals of the experiment were to: 1) test whether object saliency maps contain enough information to allow humans to match them with corresponding game scenarios, 2) test whether participants could use object saliency maps to generate reasonable explanations of the behavior of the Pacman and  
%(i.e. whether object saliency maps provide enough situation awareness) 
3) test whether object saliency maps allow participants to correctly \textit{predict} the Pacman's next action. This requires a deeper causal understanding of what may influence the Pacman in his decisions.

\begin{figure*}[!h]
\centering
\subfigure[Screen-shots]{\label{screenshot}\scalebox{0.7}{\includegraphics[width=0.65\textwidth]{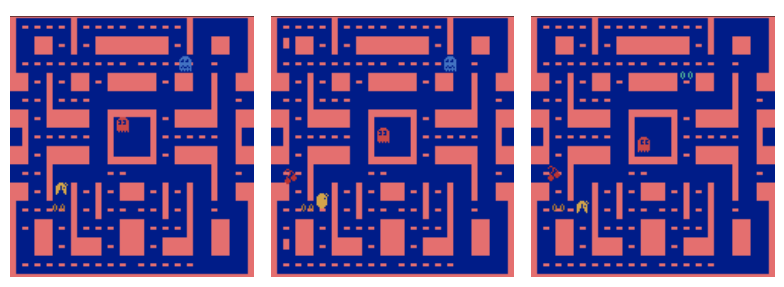}}}\hfill%
\subfigure[Object Saliency Maps]{\label{o-saliency}\scalebox{0.7}{\includegraphics[width=0.65\textwidth]{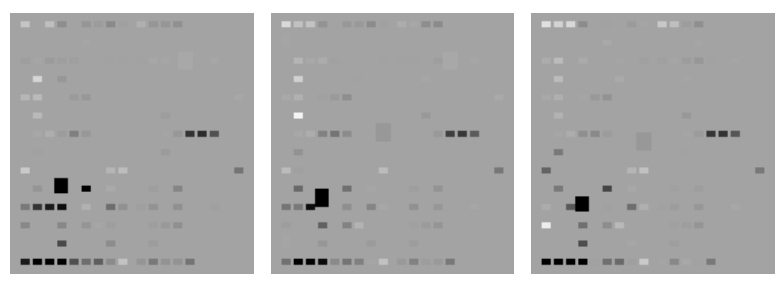}}}\hfill%
\caption{An example of the stimulus materials participants saw in the test 9 of the prediction task.  75\% participants in the screen-shot group thought Pacman would go left to eat the cherry at the left side. 60\% participants in the object saliency maps group predicted the Pacman would keep going down for the dark elements (the pellets) below.}
\label{casestudy}
\end{figure*}

Each experiment consists of two tasks: 

\textbf{Matching Task}
In each trial, the participants are shown twice, a 5-second video clip of Pacman gameplay generated by O-DDQN. During the video clip, Pacman decides and takes particular actions. The last decision made produces the crucial movement of the clip (eg Pacman moves right), with the clip ending just after the crucial movement.  
%(with no additional crucial decision having being taken up to the end of the clip). This is done so the participants have only one decision/movement to think about. 
Three frames from the object saliency map are then shown to participants (see Fig. ~\ref{o-saliency}). The center frame is the frame where the  Pacman makes the crucial decision and the other two are frames from before and after that moment. In the task, participants are asked to judge whether the saliency maps accurately represent the video they just saw. In the matching cases, the saliency maps indeed were generated from the video clip the participant saw. In the non-matching cases, the three saliency map frames were generated from a different video clip. In distractor/non-matching clips, the Pacman occupies the same area of map as in the target video, but makes different movements. This is done to avoid the case where the participants solely focus on the location of the Pacman as a matching criterion, disregarding the movements and environmental factors. 

Following the match decision, if the participants' answer is "match",  they are asked to give an explanation for the Pacman's movements based on the video and saliency maps. In other words, participants are asked to provide a teleological explanation explaining ‘why’ Pacman acted as she did. For example, ''Pacman moved up to eat more energy pellets while avoiding the ghost coming from below.'' 
%%Participants in the object saliency  condition are asked to provide an explanation based on the saliency map including the influenced elements and the rationale of Pacman's decision making .	

The matching task consisted of 2 training trials and 20 test trials, half (10 trials) presenting matched video and saliency maps, the other half presenting non-matched pairs in a single randomly ordered sequence. Dependent variables  were correctness of matches and agreement between explanations and saliency maps.

\textbf{Prediction Task}
%The prediction task has two conditions with two different participant populations. 
In each trial, the participants are shown a video clip not used in the matching task.
%from a different set than the ones for the matching task. 
Each clip ends at the point where the Pacman must choose a crucial move. The participants are divided equally into two experimental conditions.  In the 
screen-shot condition, after the video clip, participants see 3 actual screen-shots from the video ending before the crucial move is taken. In the object saliency map condition, the participants see three object saliency map frames (corresponding to the screen-shot frames) 
%that participants in the screen-shot condition saw) 
after viewing the video clip (see Fig. ~\ref{casestudy}). At the decision point in the third frame Pacman's choices 
%must choose a movement from the available choices 
(up, down, left, right) may be limited by barriers indicated on the response forms. Participants are asked to predict Pacman's movement among the feasible directions based on the three previous frames (screenshots or saliency maps), and then give an explanation for their prediction which includes their judgment as to which elements of the game influenced the Pacman's decision (indicating these elements by circling them on a hardcopy of the screenshot or saliency map), and explain why Pacman made that decision. 

\begin{figure*}[!h]

\begin{minipage}[t]{0.48\textwidth}

\includegraphics[width=0.9\textwidth]{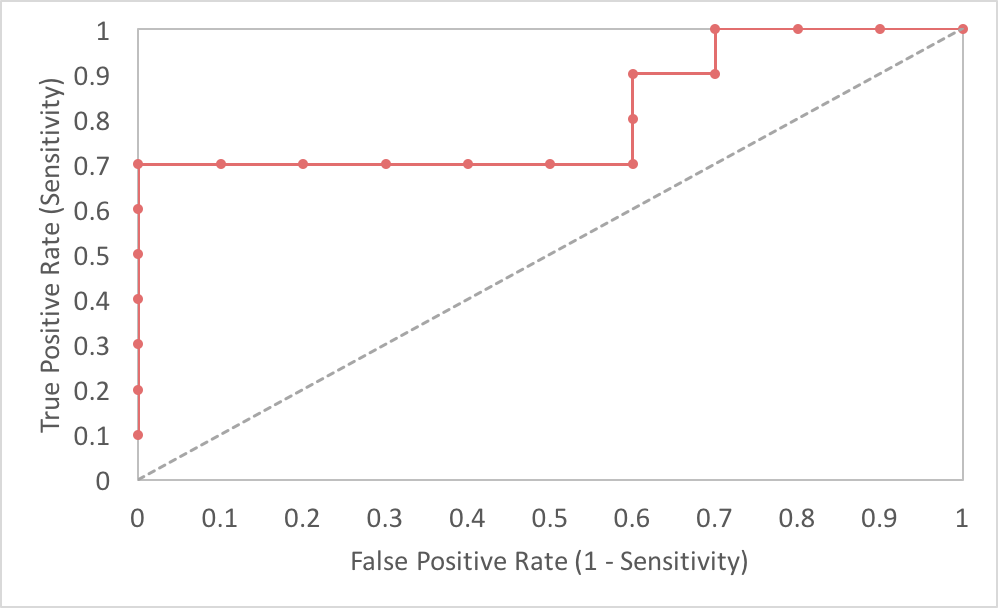}\hfill%
\caption{ROC curve of the matching task, AUC = 0.81.}
\label{roc}
\end{minipage}
\begin{minipage}[t]{0.51\textwidth}
\centering
\includegraphics[width=0.9\textwidth]{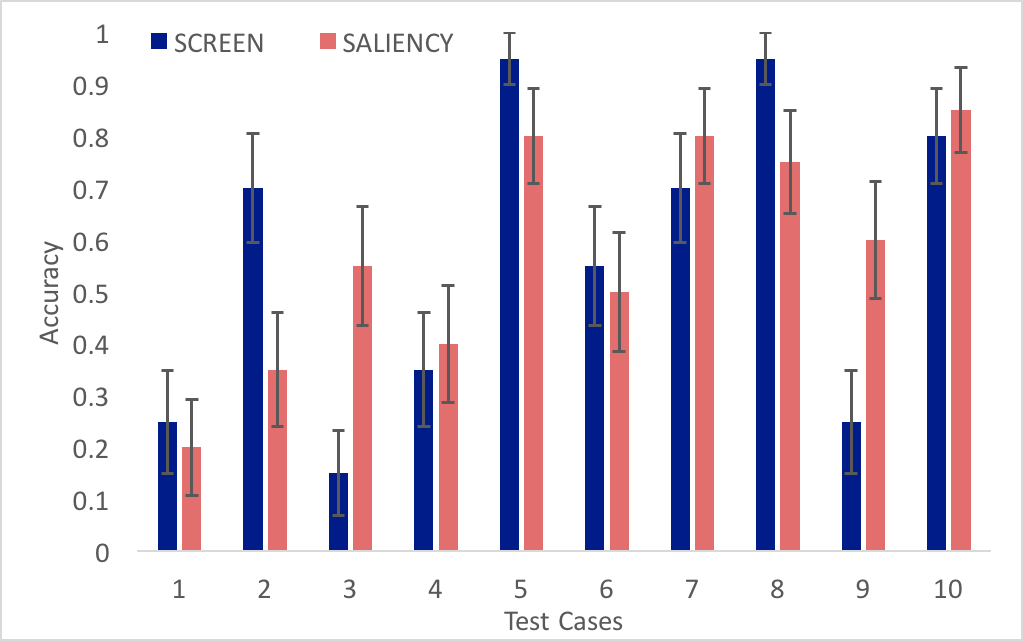}\hfill%
\caption{The mean accuracy of participants in each test cases of the prediction task. Error bars are one Standard Error from Means.}
\label{prediction-result}
\end{minipage}
\end{figure*}

The prediction task consisted of 2 training trials and 10 test trials. Each participant was assigned to either the screenshot group or the saliency map group. Dependent variables include whether predictions were correct, and whether explanations were consistent with the saliency maps.  

\textbf{Method}
40 participants were recruited from the University of Pittsburgh. %Data was collected in groups. 
The video clips and frames were presented through a projector, and participants were asked to write their answers down on answer sheets. The answer sheet included frames of each trial for participants to mark the elements they believe influenced Pacman's decision and space to explain that decision. 

{\textbf{Results}
The average matching accuracy of the participants was $61.0\%$ ($SD = 14.0\%$). A learning effect was found with participants having higher accuracy ($65.5\%$) in the last half of the trials than the first half ($56.5\%$) ($t(39) = 3.10, p = 0.04$). Comparing hit and false alarm rates, participants reported more "matches" when the video and image stimulus matched ($t(18)=2.91,p < 0.001$). If the 40 participants are treated as a binary classifier and the percentage of their answers as an output score, a receiver operating characteristic (ROC) curve ~\cite{Fawcett2006} can be plotted for true positive rates versus false positive rates across a range of threshold parameters (as Fig.~\ref{roc} shows). The area under the curve is 0.81 which indicates a good classification between matching and non-matching situations. In summary, human participants were able to link the object saliency maps with the game scenarios. %\hl{(huao: I draw the ROC curve in the way that each point means I set the judgment threshold at the percentage of answers of a certain instance. For example, for a certain instance that 95\% participants answer "match", I set the threshold at 95\%. For any other instances with a higher percentage of "match" answer, the classification output are marked as "match", and those have a lower percentage are marked as "non-match". Based on the result generated at such threshold, I calculate the TPR and FPR and draw a point on the plot. Then repeat the same thing to all other instances to finish the curve. Actually I am not 100\% sure whether the method is correct or not, since it's more common in the evaluation of machine learning classifiers.)}

For the more difficult prediction task, there was no significant difference in accuracy between the object saliency map group ($58.0\%\pm12.8\%$) and the control group($56.5\%\pm10.4\%$).  %Comparison with an exact Poisson Binomial distribution showed neither group differed significantly (p>.05) from chance in their choice of directions although the trend was toward agreement.  
However, the main effect of trials ($F(9,342) = 11.18$, $p < 0.001$) and the interaction between trials and groups ($F(9,342) = 2.72$, $p = 0.005$) were both highly significant suggesting that characteristics of the trials had a strong influence on performance. Thus we conducted a simple effect analysis to examine differences among the 10 test scenarios (see Fig.~\ref{prediction-result}). Results show that the screen-shot group has high predictive accuracy in test 2 ($p = 0.027$), while the object saliency map group has higher accuracy in tests 3 and 9 ($p = 0.007, p = 0.025$). Those three trials can help provide a deeper insight into the mechanism of how object saliency maps could help humans understand Pacman's learned behavior.}

{\textbf{Discussion}
The result of our human experiments show that object saliency maps can be linked to  corresponding game scenarios by participants, a prerequisite if they are to provide explanations of behavior.  Object saliency maps and screen shots proved equally helpful to humans in predicting DRLN's behaviors. 

For the prediction task the group viewing screen shots had access to rich contextual information including obstacles in the environment and the identity of objects making the association between the frames they viewed and rules of the game explicit.  The object saliency participants, by contrast, lacked clear identity of objects or environmental features but viewed the valence of objects affecting the decision (via the object shading).  That these complementary representations led to equal performance suggest they are both of value and deserve closer attention.  The large differences in performance found between trials, however, suggests that examining the conditions under which each provides more accurate prediction could lead to better understanding and use of object saliency maps in explanation.

Test 9 provides a good example (see Fig.~\ref{casestudy}). The  Pacman goes down and faces a dilemma whether to turn left or keep going down. 60\% participants who saw the object saliency maps predicted  Pacman would continue going down, and objects circled and explanations focused on the dark elements or dots below. In contrast, 75\% participants in the screen-shot group predicted Pacman would go left, and all except one of their explanations mentioned the cherry at the left side. In the scenarios generated by O-DDQN, the Pacman did go down for the dots. Test 9 is a typical case in which there are multiple influencing elements and it is hard for humans to predict Pacman's behavior based on information from the game screen and their own knowledge of the rules and ideas about gameplay. However, displaying object saliency enables us to directly identify those objects affecting the program's decision. In other situations when the Pacman may make what we judge to be suboptimal choices (e.g. the Pacman chose a wrong direction and was eaten by a ghost), an object saliency map could be crucial to helping users and system developers understand some of the rationale behind such behaviors.  

In watching a program such as O-DDQN play Pacman it is tempting to interpret its actions in human terms of seeking cookies and avoiding ghosts.  This in fact is what we asked participants in the screen-shot group to do.  O-DDQN, however, has no knowledge of game rules and has simply learned to maximize its reward.  In many cases due to the reward structure of the game its decisions may happen to match our own and we can attribute teleological causes, however, in others the strangeness of its decision making is revealed and we must turn to tools such as the saliency map for help. Such tools offer a better chance to improve the model when the agent executes unexpected or abnormal behaviors (e.g. debugging and testing of the DRLN).  Alternately, the agent's policies could be examined to identify why they may have been learned and what benefits they might confer leading to a deeper understanding of the domain and improved decision making.

Performance of the object saliency group on the prediction task may have suffered due to insufficient training and limitations inherent in group testing.   
%and the limitation of test sampling. Considering the learning effect and the naive background of our participants, 
We believe that a more comprehensible tutorial and longer training section might lead to better understanding of the object saliency map and improved performance in both tasks. As Fig.~\ref{prediction-result} shows, the performance pattern of participants in the prediction task depends crucially on situations. If those readily predicted from screen shots could be discriminated from those not amenable to naive explanation, saliency maps could be tested and used under more favorable conditions.

%More detailed examples for how object saliency maps can help explain the decisions made by each model can be found in the Appendix.~\ref{appendix:obj}. These prove that the object saliency maps can be used to visually explain why a model choose a certain action. 

%\begin{figure*}[h]
%\centering
%\subfigure[Current State]{\label{cur1}\includegraphics[width=0.3\textwidth]{ex-ori-1.png}}\hfill%
%\subfigure[DDQN object saliency map]{\label{DDQN1}\includegraphics[width=0.3\textwidth]{ex-DDQN-1.png}}\hfill%
%\subfigure[O-DDQN object saliency map]{\label{O-DDQN1}\includegraphics[width=0.3\textwidth]{ex-O-DDQN-1.png}}
%\caption{From left to right are the current state, the object saliency map produced by the DDQN model and the object saliency map produced by the O-DDQN model. The DDQN model chooses to go \textbf{left} while the O-DDQN model chooses to go \textbf{right} in this situation.}
%\label{obj-saliency-1}
%\end{figure*}

\section{Conclusion and Future Work}
In this paper, we developed techniques for integrating object recognition into Deep Reinforcement Learning models. Additionally we developed a technique for computing object-based saliencey maps. We evaluated the utility of this technique for visualization of agent decisions in the Ms Pacman game and via human experiments. 
%show that by incorporating object features, we can improve the performance of deep reinforcement learning models by a non-trivial margin. We also proposed object saliency maps for visually explaining the actions taken by deep reinforcement learning agents. 
One interesting future direction is how to use object saliency maps as a basis to automatically produce human intelligible explanations in natural language, such as ``I chose to go right to avoid the ghost''. 
Another direction is to test the ability of object features in a more realistic situation. For example, how to incorporate object features to improve the performance of self-driving cars.

\section{Acknowledgement}
This research was supported by awards W911NF-13-1-0416 and FA9550-15-1-0442. 

\clearpage

%\newpage
\bibliography{references}
\bibliographystyle{aaai}
\label{sect:bib}

\end{document}